\documentclass[11pt]{article}

\usepackage{acl}

\usepackage{times}
\usepackage{latexsym}
\usepackage[T1]{fontenc}
\usepackage[utf8]{inputenc}
\usepackage{microtype}
\usepackage{inconsolata}
\usepackage{graphicx}
\usepackage{booktabs}
\usepackage{multirow}
\usepackage{float}
\usepackage{enumitem}
\usepackage{CJKutf8}

\title{NAVER LABS System Re-implementation for the\\IWSLT 2026 Instruction-Following Task}

\author{
  Anand Kamble \\
  Florida State University \\
  \texttt{amk23j@fsu.edu}
  \And
  Aniket Tathe \\
  University of Illinois Urbana-Champaign \\
  \texttt{atathe2@illinois.edu}
}

\begin{document}
\maketitle

\begin{abstract}
We re-implement the NAVER LABS IWSLT 2025 instruction-following pipeline~\citep{lee-etal-2025-naver} for the IWSLT 2026 Shared Task (constrained condition, short audio track), adapting it to the mandated components: SeamlessM4T-v2-large~\citep{seamlessm4t} as the speech encoder and Qwen3-4B-Instruct~\citep{qwen3} as the LLM backbone.
The three-stage approach---projector alignment, text-only LoRA pre-training, and multimodal merging---is preserved from the original design.
We additionally construct 100k synthetic instruction-following examples across ten speech-centric task types (10k per task) from the provided corpora, suitable for further Stage~3 fine-tuning.
Our primary model achieves COMET 0.781 on EN--ZH speech translation and BERTScore-F1 0.346 on English SQA on the MCIF benchmark.
Code, training scripts, and generated data will be released on our GitHub.
\end{abstract}

\section{Introduction}
\label{sec:intro}

Multimodal speech LLMs such as SALMONN~\citep{salmonn}, Qwen-Audio~\citep{qwen-audio}, SpeechGPT~\citep{speechgpt}, and WavLLM~\citep{wavllm} couple a frozen speech encoder with an instruction-tuned LLM~\citep{instructgpt} via a lightweight connector, enabling flexible multi-task inference through natural language prompts.
The IWSLT 2026 Instruction-Following Shared Task~\citep{iwslt2026} formalizes this paradigm with the MCIF benchmark, evaluating unified models on ASR, multilingual ST (EN$\rightarrow$\{DE,IT,ZH\}), and SQA.
The NAVER LABS 2025 system~\citep{lee-etal-2025-naver} demonstrated a competitive three-stage pipeline in the IWSLT 2025 constrained setting~\citep{iwslt2025}, but was not publicly released.

We provide the first open-source re-implementation, adapted to the IWSLT 2026 constraints (SeamlessM4T-v2-large encoder, Qwen3-4B-Instruct LLM---replacing the LLaMA-3.1-8B~\citep{llama3} backbone used in 2025).
We further construct 100k synthetic instruction-following examples across ten speech-centric task types (Section~\ref{sec:synth}) and ablate LoRA rank and learning rate configurations for Stage~2 text pre-training.

\section{Task and Data}
\label{sec:task}

\paragraph{Shared Task.}
We participate in the \textbf{constrained condition}, short audio track~\citep{iwslt2026}.
Evaluation on MCIF uses WER ($\downarrow$) for ASR, COMET~\citep{comet} ($\uparrow$) for ST, and BERTScore-F1~\citep{bertscore} ($\uparrow$) for SQA.
Task instructions follow the natural-language prompt format of~\citet{lee-etal-2025-naver}.

\paragraph{Training Corpora.}
Core speech data is from CoVoST~2~\citep{covost2} and EuroParlST~\citep{europarlst} (ASR/ST) and LibriSQA~\citep{librisqa} (SQA); multilingual SQA pairs in DE, IT, ZH are obtained by machine-translating LibriSQA via SeamlessM4T-v2.
Stage~3 additionally draws on NUTSHELL~\citep{nutshell} for speech summarization and YTSeg~\citep{ytseg} for audio chapter detection.
Table~\ref{tab:data} summarizes corpora per stage.

\begin{table}[t]
\centering
\small
\begin{tabular}{llll}
\toprule
\textbf{Dataset} & \textbf{Task} & \textbf{Lang.} & \textbf{Stage(s)} \\
\midrule
CoVoST 2   & ASR, ST   & EN$\rightarrow$\{DE,IT,ZH\} & 1, 2$^*$, 3 \\
EuroParlST & ASR, ST   & EN$\rightarrow$\{DE,IT,ZH\} & 1, 2$^*$, 3 \\
LibriSQA   & SQA       & EN (+ DE,IT,ZH$^\ddagger$)  & 1$^\dagger$, 2$^*$, 3 \\
NUTSHELL   & S2TSum    & EN & 3 \\
YTSeg      & AChap     & EN & 3 \\
\bottomrule
\end{tabular}
\caption{Training corpora per stage. $^*$Text-only. $^\dagger$A.2 variant only. $^\ddagger$Machine-translated.}
\label{tab:data}
\end{table}

\begin{figure*}[t]
\centering
\includegraphics[width=0.95\textwidth]{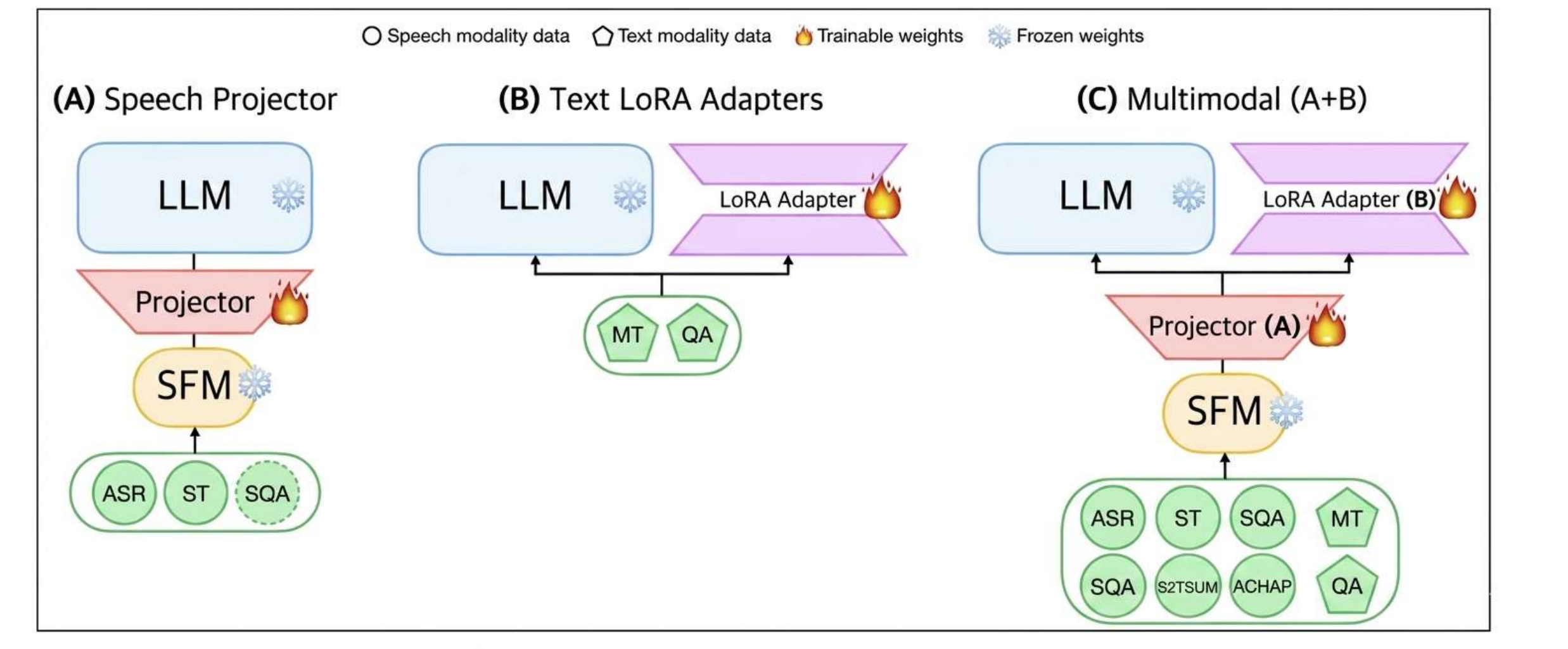}
\caption{Three-stage training pipeline. Frozen modules: dashed border. Trainable: solid. Stage~3 jointly fine-tunes both projector and LoRA adapters.}
\label{fig:pipeline}
\end{figure*}

\subsection{Synthetic Instruction-Following Data}
\label{sec:synth}

We construct 100k synthetic examples (10k per task) from the provided corpora using open-weight Gemma models~\citep{gemma}.
Seven text-grounded tasks are generated by Gemma-4-31B from reference transcripts: keyword extraction (T1), named entity recognition (T2), gist summarization (T3), topic labeling (T4), numeric QA (T5), and gist summarization in DE and ZH (T6--T7).
Three audio-grounded tasks (T8--T10) are generated by Gemma-4-E4B-it directly from audio: vocal style description in EN, DE, and ZH.

For example, a NER target (T2) for the transcript \textit{``The Luks family eventually moved to Pottsville, in southern Pennsylvania''} yields \texttt{PER=[Luks], LOC=[Pottsville, S.~Pennsylvania]}.
A vocal style target (T8) yields \textit{``The speaker has a measured, confident tone, speaking at a moderate pace with a clear articulation.''}
This data will be released with our code.

\section{System}
\label{sec:system}

\paragraph{Architecture.}
Our model follows~\citet{lee-etal-2025-naver} and is illustrated in Figure~\ref{fig:pipeline}.
A frozen SeamlessM4T-v2-large encoder~\citep{seamlessm4t} produces 1024-dim frame representations.
A trainable projector downsamples by 3$\times$ via frame averaging, passes through a 4-layer Transformer encoder~\citep{transformer}, and projects to the LLM hidden size.
A LoRA-adapted~\citep{lora} Qwen3-4B-Instruct~\citep{qwen3} generates the response, with speech embeddings prepended at a \texttt{<|speech|>} placeholder.

\paragraph{Stage 1 --- Projector Alignment.}
Encoder and LLM are frozen; only the projector is trained (4 epochs, lr $1\times10^{-4}$, constant, AdamW~\citep{adamw}).
\textbf{A.1} (ASR/ST): sampling 40\% ASR, 18\% ST-DE, 24\% ST-ZH, 18\% ST-IT (CoVoST~2 + EuroParlST).
\textbf{A.2} (ASR/ST/SQA): sampling 40\% ASR, 10.5\% ST-DE, 14\% ST-ZH, 10.5\% ST-IT, 25\% SQA-EN (adds LibriSQA).

\paragraph{Stage 2 --- Text-Only LoRA.}
No audio; projector frozen; LLM adapted via LoRA for 1 epoch.
Sampling: MT 60\% (20\% each DE/IT/ZH, from CoVoST~2 and EuroParlST transcripts) and QA 40\% (10\% each EN/DE/IT/ZH, LibriSQA + machine-translated).
Three configurations:
\begin{itemize}[noitemsep,topsep=2pt,leftmargin=1.2em]
  \item \textbf{V1}: rank~8, $\alpha$=16, lr $3\times10^{-4}$, attn+FF layers
  \item \textbf{V2}: rank~16, $\alpha$=32, lr $1\times10^{-5}$, cosine, all-linear
  \item \textbf{V3}: rank~32, $\alpha$=64, lr $2\times10^{-4}$, cosine, all-linear
\end{itemize}

\paragraph{Stage 3 --- Multimodal Merge.}
Both the A.1 projector and V1 LoRA adapters are fine-tuned jointly; the speech encoder remains frozen.
Sampling: 20\% ASR, 10\% each ST-\{DE,IT,ZH\}, 10\% SQA-EN, 5\% each SQA-\{DE,IT,ZH\}, 10\% S2TSum, 15\% AChap.
Each speech batch (ST, SQA) is immediately followed by a paired text-only batch (MT, QA) to prevent catastrophic forgetting.
Projector lr: $1\times10^{-5}$ (constant); LoRA lr: $3\times10^{-4}$ (cosine); 2 epochs on 4$\times$H100.

\section{Experiments}
\label{sec:experiments}

\begin{table*}[t]
\centering\small
\textbf{(a)~MCIF Benchmark Results (constrained, short audio)}\\[3pt]
\begin{tabular}{lcccccccc}
\toprule
& \textbf{ASR} & \multicolumn{3}{c}{\textbf{ST COMET} ($\uparrow$)} & \multicolumn{4}{c}{\textbf{SQA BERTScore-F1} ($\uparrow$)} \\
\cmidrule(lr){3-5} \cmidrule(lr){6-9}
\textbf{Model} & \textbf{WER} ($\downarrow$) & DE & IT & ZH & EN & DE & IT & ZH \\
\midrule
SeamlessM4T-v2-large (base) & 21.49 & 0.674 & 0.723 & 0.638 & 0.153 & 0.151 & 0.158 & 0.125 \\
A.1 Projector (ASR/ST)      & 28.90 & 0.698 & 0.727 & 0.763 & 0.186 & 0.185 & 0.186 & 0.186 \\
A.2 Projector (ASR/ST/SQA)  & 37.94 & 0.661 & 0.709 & 0.732 & 0.267 & 0.296 & 0.251 & 0.289 \\
\midrule
Stage~3 (A.1 + V1 LoRA)     & \textbf{23.49} & \textbf{0.707} & \textbf{0.749} & \textbf{0.781} & \textbf{0.346} & 0.266 & 0.265 & 0.189 \\
\bottomrule
\end{tabular}

\smallskip
\textbf{(b)~Stage-2 Text Eval (1k CoVoST-2, not MCIF-comparable)}\\[3pt]
\begin{tabular}{lccccc}
\toprule
& \multicolumn{2}{c}{\textbf{MT (DE)}} & \multicolumn{2}{c}{\textbf{MT (ZH)}} & \textbf{SQA} \\
\cmidrule(lr){2-3} \cmidrule(lr){4-5}
\textbf{Model} & BLEU & COMET & BLEU & COMET & F1 \\
\midrule
Qwen3-4B (base)    & 25.48 & 0.824 & 2.03  & 0.865 & 0.482 \\
+Stage~2 LoRA (V1) & \textbf{31.69} & \textbf{0.852} & \textbf{13.18} & \textbf{0.880} & \textbf{0.652} \\
\bottomrule
\end{tabular}

\smallskip
\textbf{(c)~Stage-2 LoRA Rank Ablation}\\[3pt]
\begin{tabular}{lccccc}
\toprule
& \multicolumn{2}{c}{\textbf{MT (DE)}} & \multicolumn{2}{c}{\textbf{MT (ZH)}} & \textbf{SQA} \\
\cmidrule(lr){2-3} \cmidrule(lr){4-5}
\textbf{Config} & BLEU & COMET & BLEU & COMET & F1 \\
\midrule
V1 (r=8,  $\alpha$=16) & 31.69 & 0.852 & 13.18 & 0.880 & \textbf{0.652} \\
V2 (r=16, $\alpha$=32) & 30.94 & 0.851 & 13.97 & 0.880 & 0.650 \\
V3 (r=32, $\alpha$=64) & \textbf{32.58} & \textbf{0.853} & \textbf{15.34} & 0.880 & 0.600 \\
\bottomrule
\end{tabular}
\caption{Experimental results. (a)~MCIF benchmark; Stage~3 is our primary system. (b)~Stage-2 text eval on 1k CoVoST~2; not MCIF-comparable. (c)~LoRA rank ablation (lr: V1=3e-4, V2=1e-5, V3=2e-4); V1 used in Stage~3.}
\label{tab:main}
\end{table*}

\paragraph{Setup.}
Stages~1--2 train on one H200 GPU; Stage~3 on 4$\times$H100 80GB (DDP).
Audio longer than 15 seconds is excluded due to memory constraints.
Main evaluation uses the official \texttt{mcif\_eval} tool; Stage~2 is separately evaluated on a 1k CoVoST~2 text subset (not MCIF-comparable).

\paragraph{Stage 1 \& 3 Results (MCIF).}
Table~\ref{tab:main}(a) shows MCIF results.
Stage~1 improves ST and SQA over the SeamlessM4T-v2-large baseline at the cost of higher WER---consistent with~\citet{lee-etal-2025-naver}.
A.2 boosts English SQA (0.267 vs.\ 0.186) but further degrades ASR.
Stage~3 recovers ASR (23.49 WER), achieves the best ST COMET across all pairs (EN--ZH: 0.781), and strongly improves English SQA (0.346).
Cross-lingual SQA remains lower due to sparse multilingual supervision.

\paragraph{Stage 2: Text Evaluation \& LoRA Ablation.}
Since Stage~2 is text-only (no audio), its results are evaluated on the 1k CoVoST~2 text subset and are not comparable to MCIF.
Table~\ref{tab:main}(b) shows the gain from LoRA (V1) over the base Qwen3-4B~\citep{bleu,sacrebleu}; Table~\ref{tab:main}(c) ablates LoRA rank.
V3 (rank~32) achieves the highest MT scores; V1 (rank~8) yields the best SQA F1 and is selected for Stage~3.

\section{Conclusion}
\label{sec:conclusion}

We re-implemented the NAVER LABS three-stage instruction-following pipeline~\citep{lee-etal-2025-naver} for the IWSLT 2026 constrained setting, adapting it to SeamlessM4T-v2-large and Qwen3-4B-Instruct.
Our Stage~3 model achieves COMET 0.781 on EN--ZH ST and BERTScore-F1 0.346 on English SQA on MCIF, with consistent improvements over projector-only baselines.
The 100k synthetic dataset released alongside our code provides a natural extension point for richer Stage~3 fine-tuning or future reinforcement learning~\citep{grpo} with task-specific rewards and LLM-as-judge criteria~\citep{prometheus}.

\section*{Limitations}

Audio longer than 15 seconds is excluded due to GPU memory constraints, potentially limiting performance on longer utterances.
Cross-lingual SQA relies on machine-translated QA pairs, which may introduce noise.
Stage~2 ablation numbers are evaluated on a text-only subset and are not directly comparable to MCIF results.



\clearpage
\bibliography{custom}

\clearpage
\appendix

\section{Synthetic Dataset: Task Definitions and Examples}
\label{app:synth}

Each of the 100k synthetic examples (10k per task) is paired with a natural-language instruction prompt; the model is trained to produce the target output.
Tasks T1--T7 are \textit{text-grounded}: Gemma-4-31B generates targets from reference transcripts.
Tasks T8--T10 are \textit{audio-grounded}: Gemma-4-E4B-it generates targets directly from speech audio.
One example per task is shown below.

\paragraph{T1 --- Keyword Extraction.}
\textit{Definition:} Extract the most salient keywords from the spoken utterance as a comma-separated list.\\
\textbf{Input:} \textit{``These data components in turn serve as the `building blocks' of data exchanges.''}\\
\textbf{Output:} data components, building blocks, data exchanges

\paragraph{T2 --- Named Entity Recognition.}
\textit{Definition:} Identify and classify named entities (PER, ORG, LOC) present in the transcript.\\
\textbf{Input:} \textit{``Saunders was born in Farnborough, Hampshire, the son of a British Army officer.''}\\
\textbf{Output:} \texttt{PER:[Saunders], ORG:[British Army], LOC:[Farnborough, Hampshire]}

\paragraph{T3 --- Gist Summarization (EN).}
\textit{Definition:} Produce a one-sentence English summary of the spoken content.\\
\textbf{Input:} \textit{``...if you don't want to look at all of the man pages, you just want to see a very short description of what a command does---type \texttt{whatis} followed by the command name...''}\\
\textbf{Output:} The \texttt{whatis} command provides a one-line description of any Unix/Linux command, e.g.\ \texttt{whatis sort} or \texttt{whatis cat}.

\paragraph{T4 --- Topic Labeling.}
\textit{Definition:} Assign a single short topic label (one to three words) to the utterance.\\
\textbf{Input:} \textit{``These data components in turn serve as the `building blocks' of data exchanges.''}\\
\textbf{Output:} data exchange

\paragraph{T5 --- Numeric QA.}
\textit{Definition:} Generate a question--answer pair where the answer is a number explicitly mentioned in the utterance.\\
\textbf{Input:} \textit{``Only one referendum has been held at the national level in Guyana.''}\\
\textbf{Output:} Q: How many national-level referendums have occurred in Guyana? \textbar\textbar\ A: 1 referendum

\paragraph{T6 --- Gist Summarization (DE).}
\textit{Definition:} Produce a one-sentence German gist summary of the spoken content.\\
\textbf{Input:} \textit{``...if you don't want to look at all of the man pages, you just want to see a very short description of what a command does...''}\\
\textbf{Output:} Der Befehl \texttt{whatis} liefert eine einzeilige Beschreibung eines Unix-Kommandos, etwa \texttt{whatis ls} oder \texttt{whatis sort}.

\paragraph{T7 --- Gist Summarization (ZH).}
\textit{Definition:} Produce a one-sentence Mandarin Chinese gist summary of the spoken content.\\
\textbf{Input:} \textit{``...if you don't want to look at all of the man pages, you just want to see a very short description of what a command does...''}\\
\textbf{Output:} \begin{CJK}{UTF8}{gbsn}使用\texttt{whatis}命令可以快速查看Linux命令的简要描述。\end{CJK}

\paragraph{T8 --- Vocal Style Description (EN).}
\textit{Definition:} Describe the speaker's vocal style, tone, and delivery in English, inferred directly from the speech audio (no transcript used).\\
\textbf{Input:} \textit{[audio: ``These data components in turn serve as the `building blocks' of data exchanges.'']}\\
\textbf{Output:} The speaker has a measured, confident tone, speaking at a moderate pace with clear articulation.

\paragraph{T9 --- Vocal Style Description (DE).}
\textit{Definition:} Describe the speaker's vocal style in German, inferred directly from the speech audio.\\
\textbf{Input:} \textit{[audio: ``These data components in turn serve as the `building blocks' of data exchanges.'']}\\
\textbf{Output:} Die Sprecherin klingt ruhig und professionell mit einem gleichmäßigen Sprechtempo und einer freundlichen, sachlichen Betonung.

\paragraph{T10 --- Vocal Style Description (ZH).}
\textit{Definition:} Describe the speaker's vocal style in Mandarin Chinese, inferred directly from the speech audio.\\
\textbf{Input:} \textit{[audio: ``These data components in turn serve as the `building blocks' of data exchanges.'']}\\
\textbf{Output:} \begin{CJK}{UTF8}{gbsn}声音听起来平稳而中速，带有平静的语调，情感色彩比较客观。\end{CJK}

\end{document}